\documentclass[10pt,twocolumn,letterpaper]{article}

\usepackage{mystyle}
\usepackage{times}
\usepackage{epsfig}
\usepackage{graphicx}
\usepackage{amsmath}
\usepackage{amssymb,xcolor}
\usepackage{tabularx}
\usepackage{booktabs}
\usepackage{arydshln}
\usepackage{subcaption}
\usepackage[ruled,vlined]{algorithm2e}

\makeatletter
\@namedef{ver@everyshi.sty}{}
\makeatother
\usepackage{tikz}

\usepackage{pgfplots}
\usepackage{multirow}
\usepackage{multicol}
\usepackage{enumitem}
\usepackage{hhline}
\usepackage{boldline,microtype}
\iccvfinalcopy

\def\etal{et al. }
\newcommand{\mname}{\textit{Positional Diffusion}\xspace}
\newcommand{\mnamenoit}{Positional Diffusion\xspace}
\def\toprule{\hline\hline}
\def\midrule{\hline}
\def\bottomrule{\hlineB{2.7}}

\usepackage[pagebackref=true,breaklinks=true,letterpaper=true,colorlinks,bookmarks=false]{hyperref}

\begin{document}

\title{\mnamenoit: Ordering Unordered Sets \\ with Diffusion Probabilistic Models} %w

\author{Francesco Giuliari$^*$ $^{1,2}$ \quad Gianluca Scarpellini$^*$ $^{1,2}$ \quad Stuart James$^{1}$ \quad Yiming Wang$^{1,3}$ \quad Alessio Del Bue$^{1}$\\
$^1$ Istituto Italiano di Tecnologia (IIT) \quad $^2$ University of Genoa \\
$^3$ Fondazione Bruno Kessler (FBK) \\
$^*$ equal contribution
}

\maketitle

\begin{abstract}
Positional reasoning is the process of ordering unsorted parts contained in a set into a consistent structure. We present \mname, a plug-and-play graph formulation with Diffusion Probabilistic Models to address positional reasoning. 
We use the forward process to map elements' positions in a set to random positions in a continuous space. \mname learns to reverse the noising process and recover the original positions through an Attention-based Graph Neural Network.
We conduct extensive experiments with benchmark datasets including two puzzle datasets, three sentence ordering datasets, and one visual storytelling dataset, demonstrating that our method outperforms long-lasting research on puzzle solving with up to $+18\%$ compared to the second-best deep learning method, and performs on par against the state-of-the-art methods on sentence ordering and visual storytelling. Our work highlights the suitability of diffusion models for ordering problems and proposes a novel formulation and method for solving various ordering tasks. Project website at \href{https://iit-pavis.github.io/Positional_Diffusion/}{https://iit-pavis.github.io/Positional\_Diffusion/}%

\end{abstract}
\section{Introduction}
\label{sec:introduction}

\begin{figure}
\includegraphics[width=\linewidth]{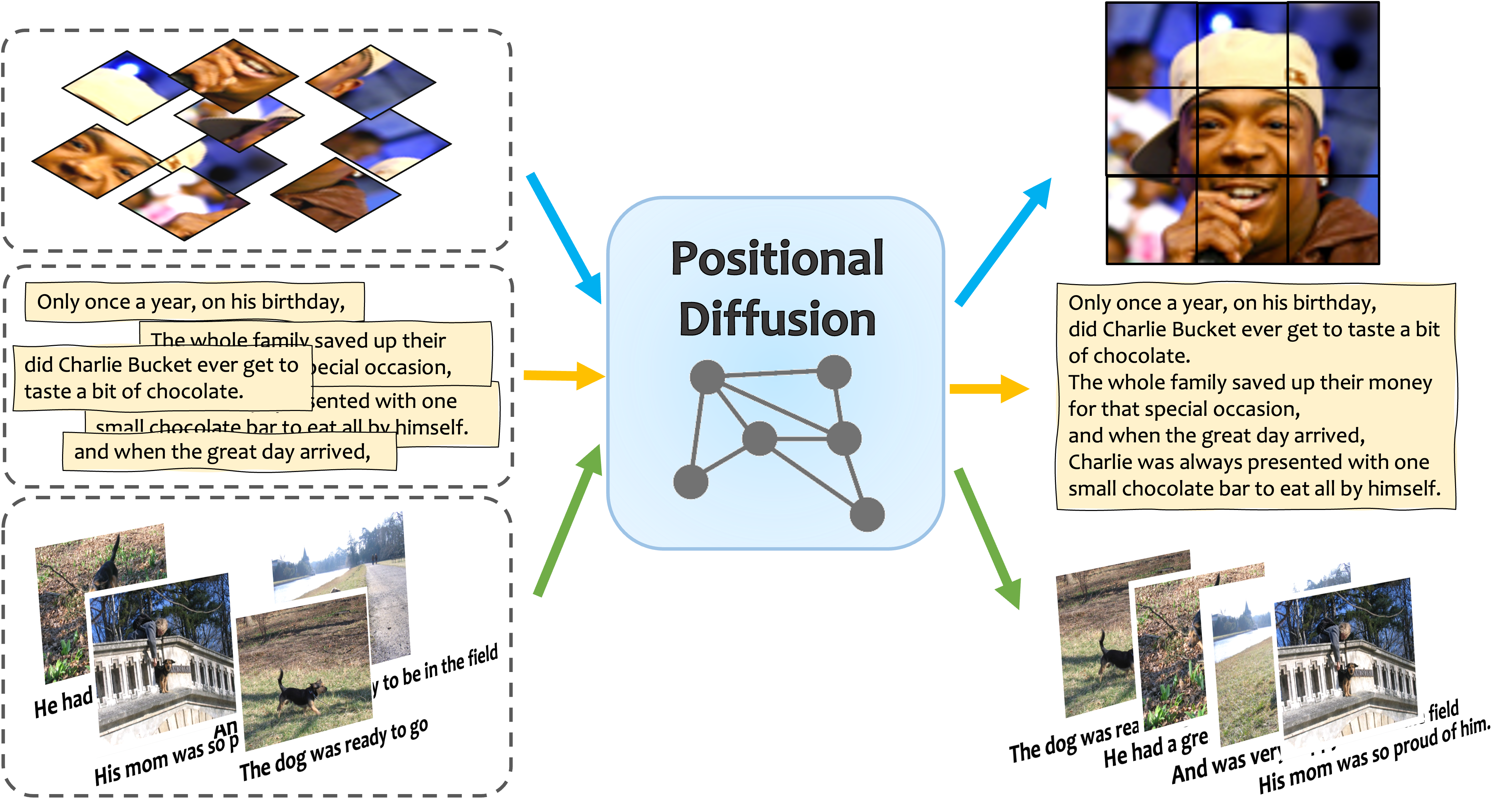}
\caption{\mname is a unified architecture based on Diffusion Probabilistic Models following a graph formulation. It can solve several ordering problems with different dimensionality and multi-modal data such as Jigsaw puzzles, sentence ordering and coherent visual storytelling.}
\label{fig:fig1} 

\end{figure}

The ability to arrange elements is a fundamental human skill that is acquired during the early stages of development and is essential for carrying out daily tasks.
Such ability is general across different tasks and researchers suggest that childhood games, such as Jigsaw puzzles, Lego\textsuperscript{\textcopyright} blocks, and crosswords play a critical role in building the foundations of reasoning over the correct arrangement of things~\cite{levine2012early, verdine2014deconstructing}. While each of these tasks is tackling a very specific problem, humans have remarkable skills in \textit{``putting an element in the correct place''} regardless of the dimensionality and the information modality of the problems, such as 1-dimensional (1D) for arranging texts or 2D for solving puzzles.
We refer to this ability as \textit{positional reasoning}, and formulate it as an \textit{ordering} problem, i.e., assigning a correct \textit{discrete} position to each element of an unordered set. 

The difficulty lies in the combinatorial nature of ordering a set of elements into a coherent (given) structure. 
A robust ordering method has to be invariant to random permutations of the input sets, while consistently providing the correct output.  
Previous solutions have been designed to be problem-specific. %
For example, methods addressing Jigsaw puzzle operates on a 2D grid by jointly optimizing similarities and permutations~\cite{zhanglearning} or by learning first an image representation complaint with the set of image tiles and then solving a standard Hungarian approach for matching the pieces~\cite{talon2022ganzzle}.
Sentence ordering is another relevant 1D ordering NLP problem where a paragraph is formed from a set of unordered sentences by exploiting pairwise similarities and attention mechanisms~\cite{gong2016end, logeswaran2018sentence, cui2018deep, yin2019graph, yin2020enhancing}.
Although all these positional reasoning problems involve finding a correct ordering of a set, their solutions are mostly customized to the data structure, position dimensionality, and contextual information.

We propose a unified model for positional reasoning, which does not require a re-design of the architecture given different input modalities or the dimensionality of the positional problem. 
We solve the ordering problem by regressing the position of each element in the set in a bounded continuous space. We then use the continuous position to retrieve the element's ordering in the set. %
Our approach is based on Diffusion Probabilistic Models (DPM) to estimate the position (and thus ordering) of each element in the set. We achieve permutation invariance by representing elements in the set as nodes of a fully connected graph. %
Using a diffusion formulation at training, we inject noise to the node positions and train an Attention-based Graph Neural Network (GNN) to learn the reverse process that recovers the correct positions. The attention mechanism aggregates relevant information from neighboring nodes given the current node features and positions. At inference, we initialize the graph with sampled positions and iteratively retrieve the correct ordinal positions by conditioning on nodes' features.

Our proposed method, named \mname, can address various problems that require ordering an arbitrary set in a plug-and-play manner.
In this paper, we demonstrate the effectiveness of our formulation and method with three fundamental tasks: \textit{i) puzzle solving}, where we compare \mname to both optimization-based and deep-learning-based methods, scoring the new state-of-the-art (SOTA) performance among all methods with a margin up to $+18\%$ compared to the second-best deep-learning method; \textit{ii) sentence ordering}, where we obtain the SOTA performance in a subset of the test datasets, including NeurIPs Abstract~\cite{logeswaran2018sentence} and Wikipedia Plots~\cite{chowdhury2021everything}; and \textit{iii) visual storytelling}, where \mname is on par with the SOTA methods on the VIST dataset~\cite{huang2016visual} without relying on an ensemble of methods or any matching algorithms.
 
To summarize, our main contributions are the following:
\begin{itemize}[noitemsep,nolistsep]
    \item We propose a novel graph formulation with DPMs to address positional reasoning. The graph formulation addresses the invariance to input set permutations while the DPMs learn to restore the positions via the noising and de-noising processes;
    \item We propose a task-agnostic method, \mname, that implements an Attention-based GNN following a DPM formulation to address positional reasoning in various tasks in a plug-and-play manner;
    \item We show without any task-specific customization, \mname can generalize and achieve SOTA or on-par performance among existing methods that are specifically designed for the tasks.
\end{itemize}

\begin{figure*}[t]
     \centering
     \includegraphics[width=\linewidth]{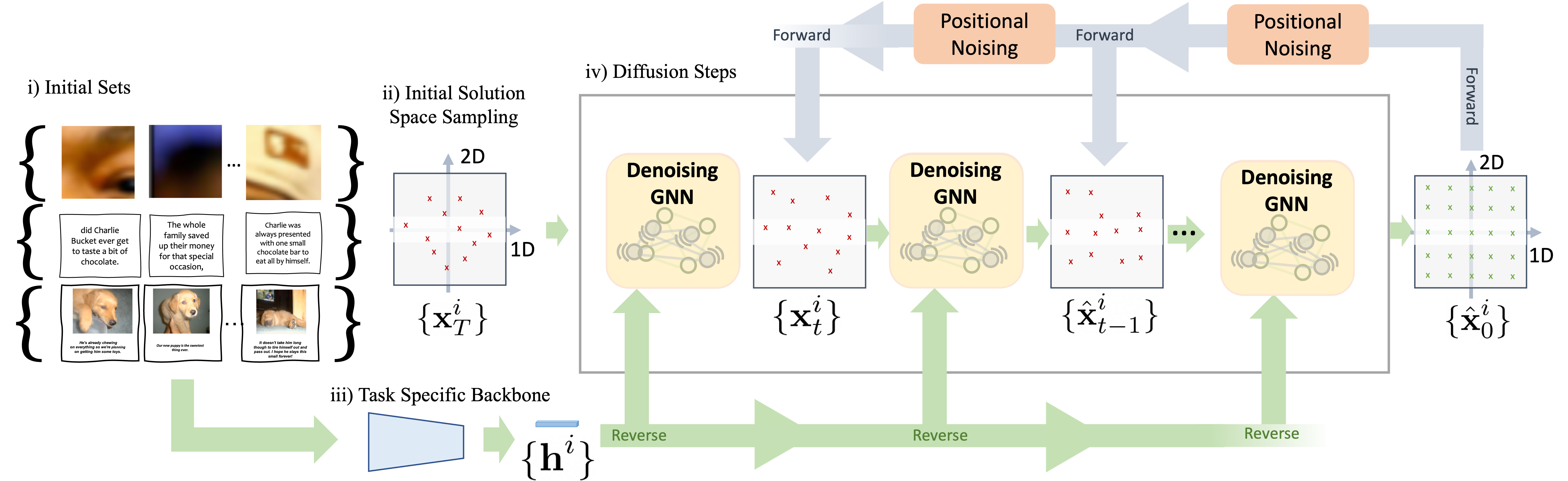}
     \caption{For each task, the input initial set (i) is a permuted version of the solution. Each element of the set is correlated with an initial sample location (ii) $\textbf{x}^i_{T}$ (in 1D or 2D) and an encoding $\textbf{h}^i$ from a task-specific backbone (iii). During Training of the diffusion steps (iv) we apply a noising process to each element position $\textbf{x}^i$ to obtain a noisy position $\textbf{x}_t^i$. We concatenate $\textbf{h}^i$ with the noisy positions $x_t^i$ to create the features and encode them as node features in a fully connected graph. We use a Graph Neural Network with Attention-based message passing to generate the less noisy positions $\textbf{x}_{t-1}^i$. During inference, for each element, we sample an initial position $\textbf{x}_T^i$ from $\mathcal{N}(0,1)$ or set it to $[0,0]$, and use \mname for the full reverse process to obtain the estimated positions $\hat{\textbf{x}}^i_0$.
     }
     \label{fig:network_arc}
 \end{figure*}

\section{Related works}
\label{sec:related}
We consider related works on recent developments of Diffusion Probabilistic Models and the SOTA methods of the three representative tasks for positional reasoning, including \textit{puzzle solving}, \textit{sentence ordering}, and \textit{visual storytelling}. 

\noindent\textbf{Diffusion Probabilistic Models.}
Diffusion Probabilistic Models (DPMs) solve the inverse problem of removing noise from a noisy data distribution \cite{sohl2015deep}. They gained popularity thanks to their impressive results on image synthesis~\cite{ho2020denoising,dhariwal2021diffusion} and their elegant probabilistic interpretation~\cite{song2020score}. Recent literature showed that DPMs' capabilities extend to the 3D space~\cite{luo2021diffusion} and unveiled promising applications for molecule generation~\cite{hoogeboom2022equivariant}. We propose a novel formulation of the forward and reverse diffusion process for coherently sorting a shuffled input by treating the problem as either $n$-dimensional vectors sampled from a Gaussian distribution.%
We are unaware of previous works proposing an extensive study on positional reasoning with diffusion models.

\noindent\textbf{Positional reasoning.}
Literature on positional reasoning is vast and assumes different connotations depending on the task and modalities involved. Our study focuses on positional reasoning as an ordering task, i.e., sorting shuffled elements into a coherent output. 

 \noindent\textbf{i)} \textit{Jigsaw Puzzles} \cite{choCVPR10probjigsaw} interested the optimization community with puzzles as a benchmark for studying image ordering with intrinsic combinatorial complexity. The most successful strategies are related to greedy approaches using hand-crafted features \cite{gallagher2012jigsaw,pomeranzCVPR11greedy} with robustness to noise and missing pieces \cite{PaikinCVPR2015} and solving thousands of pieces. For deep learning, puzzles have also been addressed as a permutation problem. Zhang \etal \cite{zhanglearning} optimizes both the cost matrix assessing pairwise relationships between pieces and the correct permutation in a bi-level optimization scheme, retaining the iterative elements of optimization methods. Alternatively, \cite{zhang2018ICLR} used puzzles as a self-supervised representation learning task, where concatenation fixed the number of solvable pieces without optimization. Talon \etal~\cite{talon2022ganzzle} overcame this by exploiting a GAN~\cite{goodfellow2020generative} and reframing the problem as an assignment problem against the generated image.

 \noindent \textbf{ii)} \textit{Sentence ordering} involves positional reasoning on textual contents, which aims to order sentences into a coherent narration. Early works solved the task by modeling local coherence using language-based features~\cite{LapataB05, barzilay2008modeling, elsner2011extending, guinaudeau2013graph}. Recent works leverage deep learning to encode sentences and retrieve the final order using pointer networks, which compare sentences in a pairwise fashion~\cite{vinyals2015pointer,gong2016end, logeswaran2018sentence, cui2018deep, yin2019graph, yin2020enhancing}. Several proposed approaches utilize attention-based pointer networks~\cite{vinyals2015pointer}, topological sorting ~\cite{prabhumoye2020topological,oh2019topic}, deep relational modules~\cite{cui2020bert}, and constraint graphs to enhance sentence representations \cite{wang2019hierarchical,zhu2021neural}. Other works also reframed the problem as a ranking problem \cite{chen2016neural}, while Chowdhury \etal \cite{chowdhury2021everything} formulated sentence ordering as a conditional text generation task using a sequence-to-sequence model~\cite{lewis2019bart}.

 \noindent \textbf{iii)} \textit{Visual storytelling } is an extension of sentence ordering with both textual and visual inputs to form a coherent visual stories~\cite{agrawal2016sort}. In this multi-modal task formulation, the input is a shuffled set of sentences and their visual representation. VIST dataset~\cite{huang2016visual} is the benchmark for this task. While Zellers \etal \cite{zellers2021merlot} constrained this task to visual ordering, where the sentences are presented in order, we maintain the original formulation for this task~\cite{agrawal2016sort} to study the capabilities of diffusion models in a multi-modal setting. 

Differently from previous literature in  computer vision, natural language processing, and multimodal learning, we interpret data shuffling as the noise injection of DPMs' forward process and exploit the reverse process of a DPM to retrieve the final position of each element, that being a sentence, a puzzle piece, or a sentence-image pair. To the best of our knowledge, our \mname is the first DPM-based solution for positional reasoning that can work with different modalities. 

\section{\mnamenoit}
\label{sec:method}
We introduce positional reasoning as a restoring process from a shuffled, unstructured data distribution in a Euclidean space $\mathbb{R}^n$, where $n$=1 for 1D problems such as sentence ordering, $n$=2 for 2D tasks such as puzzle pieces arrangement and so on.
Given an unordered set of $K$ elements %
with some task-specific features $\textbf{H} = \{\textbf{h}^1,\dots,\textbf{h}^K\}, \textbf{h}^i \in \mathbb{R}^d$, where $d$ is the dimension of the features, and with ground truth positions $\textbf{X}~=~\{\textbf{x}^1,\dots,\textbf{x}^K\}, \textbf{x}^i~\in~\mathbb{R}^n$, our network outputs an estimate set of positions $\hat{\textbf{X}}~=~\{\hat{\textbf{x}}^1,\dots,\hat{\textbf{x}}^K\}, \hat{\textbf{x}}^i~\in~\mathbb{R}^n$, that matches the real position of each element.
We encode each data point as a node in a fully connected graph, allowing each node to influence the others. Graph representations have the advantage to admit a variable number of input data in a permutation-invariant way.

Our proposed~\mname uses the DPMs formulation to iteratively restore the position of the unordered data from a randomly sampled position, combined with GNNs to work with our graph-structured data.

\subsection{Network architecture}
\label{sec:method:network}
To solve the reverse process, we train a neural network that given noisy positions $\textbf{X}_t$, features $\textbf{H}$ and a time step $t$, it outputs the noise $\epsilon_t$ that is used to calculate $\textbf{X}_{t-1}$.
Our network operates with element features $\textbf{h}^i$ that can be extracted from any pre-trained task-specific backbone.
We create a graph structure $G$ where each node represents an input point and assign $[\textbf{x}_t^i;\textbf{h}^i]^\top$ as node features.
We input this graph to our Attention-based GNN backbone, which  comprises a stack of four Graph Transformers layers~\cite{shi-graphtransformer}. Such Graph Transformers use the attention mechanism on top of a graph structure to control the amount of information that is gathered from neighboring nodes. The attention is indeed what allows the network to learn to perform positional reasoning. Fig.~\ref{fig:network_arc} summarizes our architecture.

\subsection{Forward and reverse process}
\begin{figure*}[t!]
    \centering
    \includegraphics[width=\linewidth]{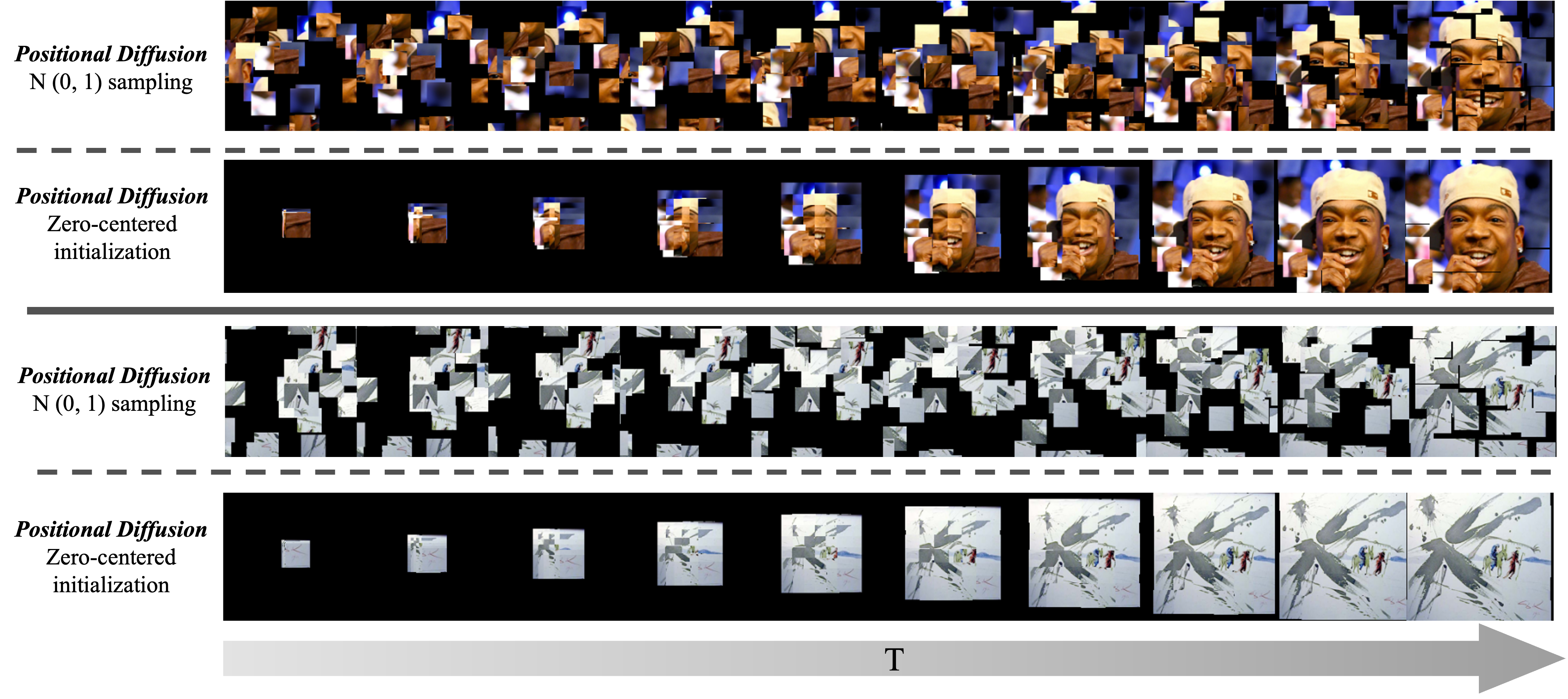}
    \caption{Visualization of the reverse process for solving puzzles of size 6x6 with both standard and zero-centered initialization. The top part shows a sample from \textit{PuzzleCelebA} while the bottom part shows the results with a sample from \textit{PuzzleWikiArt}.}
    \label{fig:qualitative_puzzles}
    \vspace{-10pt}
\end{figure*}
Building upon \cite{ho2020denoising}, we define the forward process as a fixed Markov chain that adds Gaussian noise to each input's starting position $\textbf{x}_0^k= \textbf{x}^k$ according to a Gaussian distribution. 
At timestep $t \in [0, T]$, we adopt the variance $\beta_t$ according to a linear scheduler and define $q(\textbf{x}_t | \textbf{x}_0)$ as:
\begin{equation}
    q(\textbf{x}_t^i | \textbf{x}_0^i) = \mathcal{N}(\textbf{x}_t^i; \sqrt{\overline \alpha_t} \textbf{x}_0^i, (1 - \overline \alpha_t) \textbf{I}), 
\end{equation}
where $\alpha_t = 1 - \beta_t$, $\overline \alpha_t = \prod_{s=1}^{t} \alpha_s$.
Using this formulation, we can obtain a noisy position $\textbf{x}_t^k$ from $\textbf{x}_0^k$.
The reverse process retrieves the correct position for each data point using the noisy positions $\textbf{x}_t^i$, and element features $ \textbf{h}^i$.We adopt DDIM~\cite{song2020denoising} algorithm and sample $\hat{\textbf{x}}_{t-1}$ as:

\begin{align*}
    \hat{\textbf{x}}_{t-1} = &\sqrt{\overline \alpha_{t-1}}\left(\frac{\textbf{x}_t-\sqrt{1-\overline \alpha_t} \epsilon_\theta(\textbf{x}_t,\textbf{t},\textbf{h})}{\sqrt{\overline \alpha_t}}\right) \\
    &+ \sqrt{1 - \overline \alpha_{t-1} - \sigma_t^2 }\cdot \epsilon_\theta(\textbf{x}_t,\textbf{t},\textbf{h}) + \sigma_t \epsilon,
\end{align*}

\noindent where $\epsilon_\theta (\textbf{x}_t,\textbf{t},\textbf{h})$ is the estimated noise that has to be removed from $\textbf{x}_t$ to recover $\hat{\textbf{x}}_{t-1}$ and $\textbf{t}$ is a learned vector embedding for timestep $t$. In the formula, we omit the superscripts $i$ as the network operates on all elements simultaneously as a graph. DDIM introduces the parameter $\sigma$ to control the stochastic sampling. As the ordering tasks have only one correct arrangement, we set $\sigma = 0$ to make the sampling deterministic.

Our method is trained using the simple loss for diffusion models introduced in~\cite{ho2020denoising}:
\begin{equation*}
    L_{\text{simple}}(\theta) = \mathbb{E}_{t,\textbf{x}_0,\epsilon} [\| \epsilon - \epsilon_\theta(\underbrace{\sqrt{\overline \alpha_t} \textbf{x}_0 + \sqrt{1-\overline \alpha_t}\epsilon}_{\textbf{x}_t}  ,\textbf{t},\textbf{h})  \| ].
\end{equation*}

We calculate $\textbf{x}_t$ in closed form from $\textbf{x}_0$, using the reparametrization trick with the noise vector $\epsilon$. The network learns to minimize the Mean Squared Error between $\epsilon$ and the output $\hat{\epsilon} = \epsilon_\theta(\textbf{x}_t,\textbf{t},\textbf{h})$.

\subsection{Zero-centered initialization}

\label{sec:zero_sampling}
In generative diffusion models, the initial $\textbf{X}_T$ used during the reverse process is sampled from $\mathcal{N}(0,1)$. In standard image generation tasks, this noise introduces stochasticity to synthesize different images. Differently, the solution in positional reasoning is the true final arrangement, thus such arrangement should only be influenced by the input features $\textbf{H}$ and not by the initial $\textbf{X}_T$.  
We found that by setting $\textbf{x}_T=\textbf{0}$, which is the mean of normal distribution, the network achieves more stable results as verified in the experimental section. %
The effect of different sampling on the final position can be observed in Fig.~\ref{fig:qualitative_puzzles}.
The rearrangement is more precise when $\textbf{x}_T = \textbf{0}$. A quantitative comparison is reported for puzzles, in Tab.~\ref{tab:puzzle}. We use zero-centered initialization throughout the experiments.

\begin{table}[t!]
    \centering
    \resizebox{\linewidth}{!}{
    \begin{tabular}{c|c|c|c|c|c}
    \toprule
             \multirow{2}{*}{\textsc{Task}} & \textbf{Position} & \textbf{Data} & \textbf{Feature} & \multicolumn{2}{c}{\textbf{Trainable Parameters}} \\
             & \textbf{Dim.} &\textbf{Modality} & \textbf{Backbone(s)} & \textbf{Backbone} & \textbf{GNN} \\
         \midrule
         &&&&&\\[-.3cm] 
         Puzzle solving& 2D & RGB & EfficientNet \cite{tan2019efficientnet} &             
          6.8 M & \multirow{4}{*}{3.2 M} \\[.2cm] 
         
         Sentences ordering& 1D & Text & BART \cite{lewis2019bart} &  28.2 M$^\dagger$  &\\[.2cm] 
         \multirow{2}{*}{Visual storytelling} & \multirow{2}{*}{1D} & RGB & EfficientNet \cite{tan2019efficientnet} & \multirow{2}{*}{31.8 M$^\dagger$} &\\
         &&\& Text&\& BART \cite{lewis2019bart}&&\\
\bottomrule
        \end{tabular}}
    \caption{We show the different dimensionality, modality, and number of parameters for each of our downstream tasks. Our \mname shares the same structure across tasks.  $^\dagger$We report the parameters of the trainable Transformer built on top of the frozen BART model (425 M).}
    \label{tab:summary}
\end{table}
\begin{table*}[t]
    \scriptsize
    \centering
    \begin{tabularx}{\linewidth}{X  |c@{\hskip 0.3in}  c@{\hskip 0.3in} c@{\hskip 0.3in} c@{\hskip 0.2in} |c@{\hskip 0.3in}  c@{\hskip 0.3in}  c@{\hskip 0.3in} c@{\hskip 0.1in}}
    \toprule
    \textsc{Dataset}  & \multicolumn{4}{c|}{\textbf{PuzzleCelebA}} & \multicolumn{4}{c}{\textbf{PuzzleWikiArts}} \\
      \cline{2-9}
    \textbf{} & \textbf{6x6} & \textbf{8x8} & \textbf{10x10} & \textbf{12x12} & \textbf{6x6} & \textbf{8x8} & \textbf{10x10} & \textbf{12x12} \\
    \midrule
    
    Paikin and Tal~\cite{PaikinCVPR2015}& 99.12 & \textbf{98.67} & 98.39 & 96.51  & 98.03 & 97.35 & 95.31 & 90.52 \\
    Pomeranz et al.~\cite{pomeranzCVPR11greedy} & 84.59  & 79.43  & 74.80  &  66.43  & 79.23  & 72.64  & 67.70  & 62.13 \\
    Gallagher ~\cite{gallagher2012jigsaw} &  98.55  &  97.04  & 95.49  & 93.13  & 88.77  & 82.28  & 77.17  & 73.40  \\
    \midrule
    PO-LA~\cite{zhang2018ICLR}$^\dagger$  & 71.96  & 50.12   & 38.05  & -& 12.19  & 5.77  & 3.28  & -\\

    Ganzzle~\cite{talon2022ganzzle} & 72.18 & 53.26 & 32.84 & 12.94 & 13.48 & 6.93 & 4.10 & 2.58  \\

    \hdashline
    Transformer~\cite{vaswani2017attention} & 99.60 & 95.20 & 98.62 & 96.55 & 98.52 & 95.30 & 88.76 & 75.84\\
    \hdashline

    \textbf{\mname} - $\mathcal{N}(0,1)$ sampling & 99.72 & 96.78 & 99.28 & 98.55 & 98.52 & 97.15 & 94.34 & 90.26  \\
    \textbf{\mname} - Zero-centered initialization & \textbf{99.77} & 97.53 & \textbf{99.37} & \textbf{98.88} & \textbf{99.12} & \textbf{98.27} & \textbf{96.28} & \textbf{93.26}  \\
    
    \bottomrule
    \end{tabularx}
    \centering
    \caption{Results on puzzle solving in terms of the \emph{Direct Comparison Metric}, evaluated with \textit{PuzzleCelebA} and \textit{PuzzleWikiArts}. \textbf{Best}. $^\dagger$Trained on individual puzzle sizes.}
    \label{tab:puzzle}
    \vspace{-10pt}
\end{table*}

\section{Experiment}
\label{sec:experiments}
We evaluate \mname on three tasks that require positional reasoning with input data of different modalities: \textit{i)} puzzle solving operates with visual data to order the shuffled image patches into a complete image; \textit{ii)} sentence ordering operates with textual data that aims to order the shuffled sentences to form a complete and reasonable paragraph; and \textit{iii)} visual storytelling operates with textual and visual data and requires to order sentence-image pairs into a coherent story. Tab.~\ref{tab:summary} summarizes the experimental settings.
The following sections introduce the detailed experimental setup for each task regarding the evaluation protocols, performance metrics, and comparisons. We present more qualitative results in the Supplementary Materials.

\subsection{Puzzle solving}
\label{sec:exp:puzzle}

\begin{figure*}
     \centering
     \hfill
     \begin{subfigure}[b]{0.3\linewidth}
         \centering
         \includegraphics[width=\linewidth]{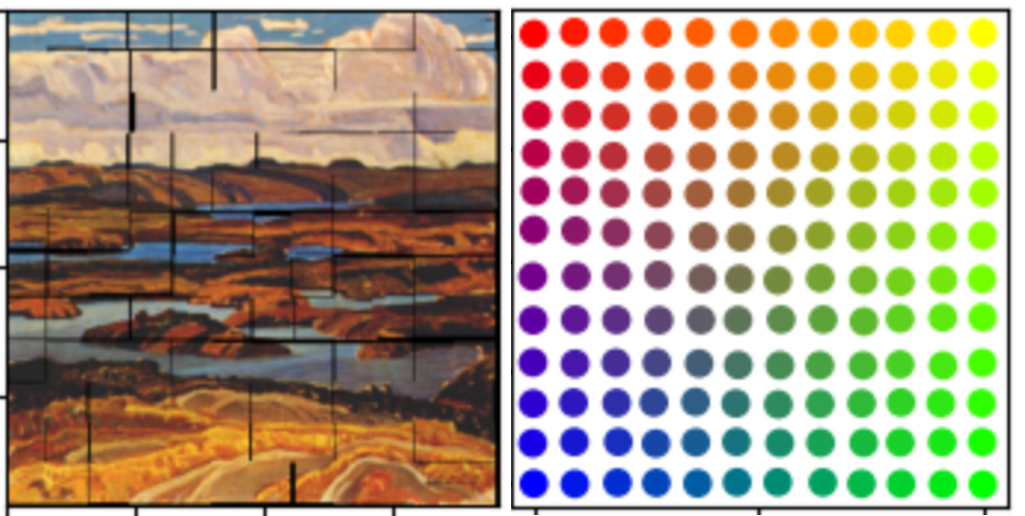}
         \caption{}
     \end{subfigure}
     \hfill
     \begin{subfigure}[b]{0.3\linewidth}
         \centering
         \includegraphics[width=\linewidth]{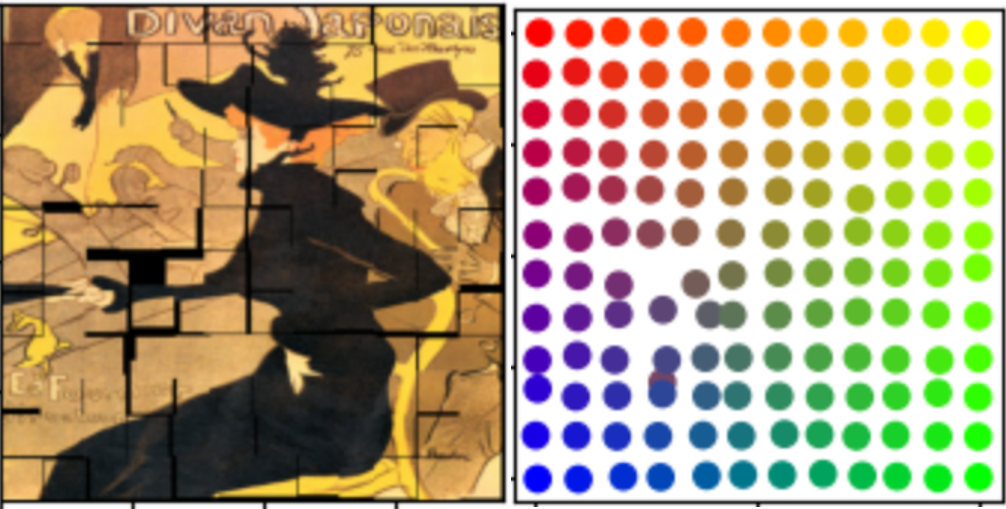}
         \caption{}
     \end{subfigure}
     \hfill
     \begin{subfigure}[b]{0.3\linewidth}
         \centering
         \includegraphics[width=\linewidth]{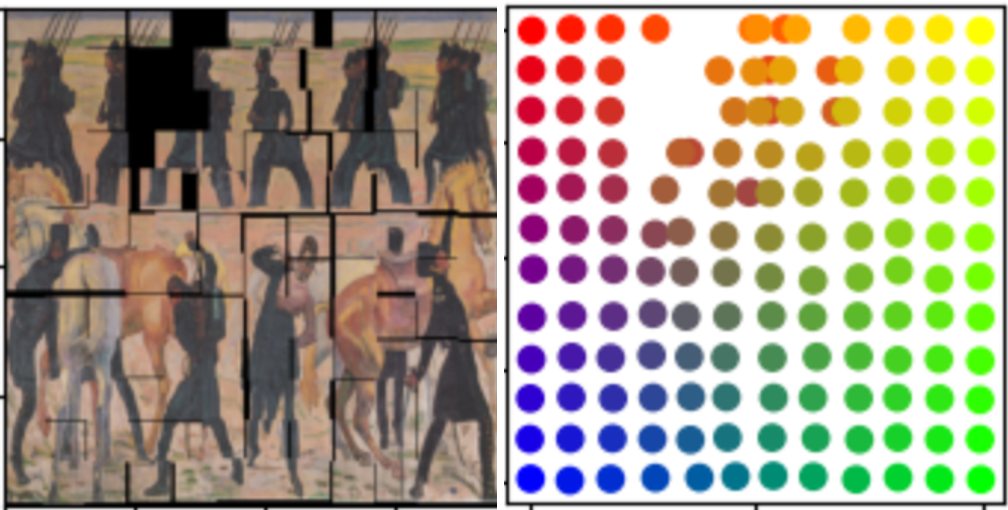}
         \caption{}
     \end{subfigure}
     \hfill
        \caption{Results of 12x12 puzzles solved with \mname on \emph{PuzzleWikiArt}. The predicted positions of patches are visualized next to the resulting puzzles. The positions are color-coded based on the patch position in the original image, from the top-left \textit{red} to the bottom-right \textit{green}. (a) The network predicts the positions of all elements correctly. (b) The network predicts the positions of the patches with slight noises. In this case, the assignment process maps the noisy positions to their correct slot. (c) The network wrongly predicts the positions of some patches in a local region. These positions cannot be recovered by the assignment process.}
        \label{fig:qualitative_puzzle_positions}
        \vspace{-10pt}
\end{figure*}

We follow the experimental setup in Ganzzle~\cite{talon2022ganzzle} and report the results of \mname in comparison to optimization-based and deep learning-based methods on \textit{PuzzleCelebA} and \textit{PuzzleWikiArts}. 
These two datasets feature a large number of images which allows for method training with deep learning, while other puzzle datasets typically only contain $\leq100$ images.
\begin{itemize}[noitemsep,nolistsep,leftmargin=*]
    \item \textit{PuzzleCelebA} is based on CelebA-HQ~\cite{CelebAMask-HQ} which contains 30K images of celebrities in High Definition (HD). The images are cropped and positioned to show only centered faces. This is an easier dataset for puzzle solving as the images share a consistent global structure.
    \item \textit{PuzzleWikiArts} is based on WikiArts~\cite{artgan2018}, and contains 63K images of paintings in HD. This dataset contains paintings with very different content and artistic styles. It represents a more challenging dataset for puzzle solving as the paintings do not have a common pattern as in PuzzleCelebA (i.e. portraits).
\end{itemize}
For both datasets, we test with puzzles of sizes 6x6, 8x8, 10x10, and 12x12.
As the puzzle size increases, the problem becomes more difficult, as the permutations increase and each piece contains less discriminative information.

\noindent \textbf{Evaluation Metrics.} We evaluate the performance of \mname using the \emph{Direct Comparison Metric}~\cite{choCVPR10probjigsaw}, a percentage that indicates the number of correctly ordered pieces over the full test set. 
A higher score indicates better performance.

\noindent \textbf{Implementation Details.} We divide an image in $\emph{n}\times \emph{n}$ patches, resulting a total of $\mathit{K}=n^2$ elements. We divide a 2D space with a range of (-1,-1), (1,1) into a grid of $\emph{n} \times \emph{n}$ cells. 
We use the centers of the cells as starting positions $\textbf{X}$ for the patches.
The input data for puzzle solving are the pixel values for each patch, resized to 32x32. We use EfficientNet~\cite{tan2019efficientnet} as the task-specific backbone to extract the patch visual features $\textbf{h}^i$ and we train the diffusion model with $T=300$ and sample it with inference ratio $r=10$.
We train a single model with all puzzle sizes simultaneously. 

At inference, we arrange the patches by mapping each estimated patch position $\hat{\textbf{x}}^i$ to a cell in the grid.
We measure the distance between each patch position and cells' centers, and assign each patch to its closest cell, mapping each cell to at most one patch. By using a greedy approach that prioritizes the assignment between cells-patch pairs starting from those with the lowest distance, we ensure that the most confident prediction will be assigned first, increasing the prediction robustness to noise.
\\
\noindent\textbf{Comparisons.}
We compared \mname against a set of SOTA methods for puzzle solving. 
\begin{itemize}[noitemsep,nolistsep,leftmargin=*]
    \item \textit{\textbf{Optimization methods}}~\cite{PaikinCVPR2015,pomeranzCVPR11greedy,gallagher2012jigsaw} are handcrafted methods for puzzle solving. They involve computing a compatibility score between all pairs of pieces, to predict which pieces are neighbors. %
   \item \textit{\textbf{PO-LA}} \cite{zhang2018ICLR} uses a neural network to learn a differentiable permutation invariant ordering cost between a set of patches. The learned cost function is then used to order the pieces and form the full image. 
   \item \textit{\textbf{Ganzzle}} \cite{talon2022ganzzle} employs a GAN to generate a hallucinated version of the full image from the set of pieces. Then, Ganzzle solves the puzzle as an assignment problem by matching the patch features with the features extracted from the generated image at specific locations.   
   \item  \textit{\textbf{Transformer}} is a standard Transformer-based architecture~\cite{vaswani2017attention} that predicts the positions of each piece. This method uses the self-attention mechanism to propagate relevant information between patches. 
\end{itemize}
\noindent 

We present two variants of \mname, where one uses the standard DPM random sampling from $\mathcal{N}(0,1)$, and the other uses the proposed zero-centered initialization for sampling.

Tab.~\ref{tab:puzzle} presents the results of all methods for solving puzzles of various sizes with the two datasets. 
On \textit{PuzzleCelebA}, both the Transformer baseline and our \mname outperform the previous SOTA methods on almost all puzzle sizes.
In particular, \mname scores the new SOTA performance among deep-learning methods on all puzzle sizes, with a significant improvement against the previous best-performing method Ganzzle~\cite{talon2022ganzzle}, even outperforming classical optimization approaches.
Moreover, we observe that the performance of previous deep learning methods degrades significantly with the puzzle sizes, while \mname only presents a minor degradation. 
On 12x12 puzzles, \mname achieves $98.88\%$, $+2\%$ higher than the Transformer baseline, $+86\%$ higher than Ganzzle, and $+2\%$ higher than the best optimization method. 
In general, \textit{PuzzleCelebA} is an easier dataset for puzzle solving compared to \textit{PuzzleWikiArts} as it contains well-centered faces with common global patterns. Our method can exploit the shared pattern to solve the puzzle, e.g., eyes position relative to the image. 
The top part in Fig.~\ref{fig:qualitative_puzzles} clearly shows that \mname can correctly position the eyes, mouth, and hair patches.

On \textit{PuzzleWikiArts}, we observe that all previous methods achieve worse performance on all puzzle sizes, among which the deep-learning approaches almost fail to solve the puzzles. \textit{PuzzleWikiArts} contains puzzles that are harder to solve, as they come from paintings with different pictorial styles and subjects, with little common patterns. 
Nevertheless, \mname consistently obtains the best performance among all methods, overshooting deep learning methods by a large margin on all puzzle sizes, i.e., $+1\%$ on 6x6, $+3\%$ on 8x8, $+8\%$ on 10x10, and $+18\%$ on 12x12 compared to the Transformer baseline.
In particular, \mname also outperforms the optimization-based methods, which require hand-crafted features and greedy solutions, on all puzzle sizes.
Moreover, using the same trained model, \mname with the zero-centered initialization consistently obtains better performance than using the standard DPM random sampling from $\mathcal{N}(0,1)$. The advantages can be observed from both datasets on all puzzle sizes, as shown in Fig.~\ref{fig:qualitative_puzzles}. Finally, we show in Fig.~\ref{fig:qualitative_puzzle_positions}, three examples of the puzzles solved with \mname on \emph{PuzzleWikiArt}. When the predicted positions are noisy but still close to their ground-truth positions, we can recover their correct ordering by the assignment procedure. However, when the errors are large and systematic, e.g., such as when there is a local  collapse in the prediction, the assignment procedure fails to fix the positions. 

It's important to note that the \textit{Direct Comparison Metric} does not reflect the performance in terms of solving a puzzle as a whole, as it is computed at the patch level. For example, the Transformer positioned $75.84\%$ patches correctly on \textit{PuzzleWikiArt} 12x12, but it only solved $6.64\%$ of the puzzles, while \mname with $93.26\%$ correctly positioned patches solved $69.32\%$ of puzzles.

\subsection{Sentence ordering}
\label{sec:method:sentence}
\begin{table*}[t!]
\centering
\scriptsize
\begin{tabularx}{\linewidth}{X|ccc|ccc|ccc}
\toprule
\multirow{2}{*}{\textsc{Method}} & \multicolumn{3}{c|}{\textbf{NeurIPS Abstract}} & \multicolumn{3}{c|}{\textbf{Wikipedia Movie Plots}} & \multicolumn{3}{c}{\textbf{ROCStories}}\\
\cline{2-10}
& \textbf{Acc} & \textbf{PMR} & $\tau$ & \textbf{Acc} & \textbf{PMR} & $\tau$ & \textbf{Acc} & \textbf{PMR} & $\tau$ \\
\midrule
TGCM \cite{oh2019topic} & 59.43 & 31.44 & 0.75 & - & - & - & - & - & -  \\
RankTxNet$^\ddagger$ \cite{kumar2020deep} & - & 24.13 & 0.75  & - & - & - & - & 38.02 & 0.76 \\
B-TSort$^\ddagger$ \cite{prabhumoye2020topological} & 61.48 & 32.59 & 0.81 & - & - & - & - & - & -\\
BERSON$^\ddagger$ \cite{cui2020bert} & \textit{73.87} & \underline{48.01} & \underline{0.85} & - & - & -& \underline{82.86} & \underline{68.23} & \underline{0.88} \\
\textsc{Bart} (seq2seq)$^\dagger$ \cite{chowdhury2021everything} & 64.35 & 33.69 & 0.78 & 30.01 & 18.88 & 0.59 & \textit{80.42} & \textit{63.50} & \textit{0.85}  \\
Re-BART$^\dagger$ \cite{chowdhury2021everything} & \textbf{77.41} & \textbf{57.03} & \textbf{0.89} & \textit{42.04} & \textbf{25.76} & \textbf{0.77} & \textbf{90.78} &\textbf{81.88}& \textbf{0.94} \\
BART+Transformer & 70.31 & 36.58 & \textit{0.84} & \underline{48.31} & \textit{22.51}& \textit{0.61} & 71.78 & 44.59 & 0.80\\
\hdashline
\textbf{\mname} & \underline{74.44} & \textit{45.24} & \underline{0.85} & \textbf{50.41} & \underline{25.00} & \underline{0.63} & 75.12 & 54.80 & 0.82 \\ %
\bottomrule
\end{tabularx}

\caption{Comparison of accuracy, PMR, and tau for various methods on NeurIPS abstract, Wikipedia Movie Plots, and ROCStories. \textbf{Best} / \underline{second best} / \textit{third best}. $\dagger$ requires fine-tuning BART. $\ddagger$ requires fine-tuning BERT.}
\label{tab:text-result}
\end{table*}

\begin{table}[t]
    \centering
        \resizebox{\linewidth}{!}{
		\begin{tabular}{ l c c c p{0.1cm} c c p{0.1cm} c c} 
			\toprule
			\multirow{2}{*}{\textsc{Dataset}} & \multicolumn{3}{c}{ \textbf{Split}} & & \multicolumn{2}{c}{\textbf{Length}} & & \multicolumn{2}{c}{\textbf{Tokens / sentence}} \\ %
            \hhline{~---~--~--}
			& {\textbf{Train}} & {\textbf{Dev}} & {\textbf{Test}} & & {\textbf{Max}} & {\textbf{Avg}} & & {\textbf{Max}} & {\textbf{Avg}} \\
			\hline
			\textit{NeurIPS Abstract}  & 2.4K & 0.4K & 0.4K & & 15 & 6 & & 158 & 24.4\\
			
			\textit{Wikipedia M. P.} & 27.9K & 3.5K & 3.5K & & 20 & 13.5 & & 319 & 20.4\\
            \textit{ROCStories} & 78K & 9.8k & 9.8k & & 5 & 5 & & 21 & 9.1\\
			\bottomrule
		\end{tabular}
  }
    \caption{Dataset statistics of \textit{NeurIPS Abstract}, \textit{Wikipedia Movie Plots} and \textit{ROCStories} for sentence ordering.}
    \label{tab:text-statistics}
    \vspace{-13pt}
\end{table}

For sentence ordering, we follow the experimental setup in~\cite{chowdhury2021everything} and report the results of all compared methods on three common textual datasets (dataset statistics are in Tab.~\ref{tab:text-statistics}):
\begin{itemize}[noitemsep,nolistsep,leftmargin=*]
    \item \textit{NeurIPS Abstract} is obtained from the abstracts of scientific articles featured at NeurIPS; 
    \item \textit{Wikipedia Movie Plots} is a collection of plots of popular movies that are available on Wikipedia; %
    \item \textit{ROCStories} is a collection of 5 sentences stories regarding everyday events.
\end{itemize}

\noindent \textbf{Evaluation metrics.} We quantify the sentence ordering performances with three metrics as in~\cite{chowdhury2021everything}:
\begin{itemize}[noitemsep,nolistsep,leftmargin=*]
    \item \textit{Accuracy (Acc.)} is the percentage of correctly predicted sentence positions in an input text.
\item \textit{Perfect Match Ratio (PMR)} is the percentage of the number of correctly ordered texts over the total number of texts in the test set. Differently from Acc. which is calculated over individual sentences, PMR measures if the full input text is ordered correctly.
\item \textit{Kendall's Tau} ($\tau$) measures the correlation between the ground-truth orders of sentences and the predicted ones, defined as: $\tau = 1- (2(\# \text{Inversions}){\binom{K}{2}}^{-1})$,
where $K$ is the number of sentences in an input text, and $\# \text{Inversions}$ is the number of discordant pairs.
\end{itemize}
We report the metrics averaged over the test set. The higher values of the three metrics indicate better performance of the sentence ordering methods. 

\noindent \textbf{Implementation Details.} We divide a text into a variable number $K$ sentences with shuffled orders as the input.
To assign the correct positions $\textbf{x}_0$ to each sentence, we evenly sample $K$ positions over the interval (-1,1), and assigned them to the divided sentences based on their position in the text.
The starting sentence will have the smallest position, while the ending sentence will have the largest position.
We use a frozen pre-trained BART~\cite{lewis2019bart} language model for our task-specific feature backbone, to which we added a learnable transformer encoder layer at the end. For each sentence, we prepend a $\langle bos \rangle$ token and pass the sentence to BART to obtain the token feature as the task-specific feature $\textbf{h}^i$ in \mname. We train our method with $T=300$ and sample with inference ratio $r=10$.

\noindent\textbf{Comparisons.} We conducted a comprehensive evaluation of \mname against the current best-performing methods BERSON \cite{cui2020bert}, Re-BART \cite{chowdhury2021everything}, and BART for seq2seq generation as proposed in~\cite{chowdhury2021everything}, as well as other baselines including B-TSort \cite{prabhumoye2020topological}, RankTxNet \cite{kumar2020deep}, TGCM \cite{oh2019topic}.
We also provide a baseline that is composed of pretrained BART backbone with a Transformer head.

We report the results of all methods in Tab.~\ref{tab:text-result}. 
\textit{Wikipedia Movie Plots} has the largest average number of sentences in a text, which is more than double compared to that of \textit{NeurIPS Abstract} and \textit{On ROCStories}. \mname scores the best Accuracy on \textit{Wikipedia Movie Plots}, with an improvement of $+8\%$ over the current SOTA method Re-BART, with on par performance in terms of PMR and $\tau$.  
With \textit{NeurIPS Abstract}, \mname is the second-best performing method in Accuracy and $\tau$, while Re-BART remains the best-performing method.
Finally, on \textit{ROCStories}, \mname performs worse than BEARSON and Re-BART. 
Compared to the well-structured texts in \textit{NeurIPS Abstract} and \textit{Wikipedia Movie Plots}, the logical connection among sentences in \textit{ROCStories} can be weak in some cases (as shown in Tab.~\ref{tab:result_rocstories}). This could be the main reason why \mname struggles to learn positional reasoning. %

Moreover, it is important to note that we use the frozen language model BART to extract features to train our GNN model for positional reasoning. Instead, Re-BART \cite{chowdhury2021everything} fine-tunes BART with all sentences simultaneously to predict the sequence order. In fact, the trainable parameters of \mname is 32M parameters for text ordering, which is negligible compared to Re-BART's 425M.

\begin{table}[t!]
\begin{center}
\resizebox{\linewidth}{!}{
\begin{tabular}{lcc}
\toprule
\textsc{Sentence} & \textbf{Predicted Pos.} &  \textbf{GT Pos.} \\ 
\midrule

Tom was driving to work. & (1) & (1)\\
He got pulled over by a cop. & (2) & (2)\\
The cop mentioned a busted tail light. & {\color{red}(3)} & {\color{red}(4)}\\
Tom asked why. &{\color{red}(4)} & {\color{red}(3)} \\
Tom agreed to fix it and only got a warning. & (5) & (5) \\
\bottomrule
\end{tabular}
}
\end{center}
\caption{Result of five sentences from \textit{ROCStories} ordered based on the positions predicted by \mname. On the right is the ground-truth position in the full text. We highlight in {\color{red}red} the wrongly predicted positions.}
\label{tab:result_rocstories}
\vspace{-13pt}
\end{table}

\subsection{Visual storytelling}
\label{sec:vist_experiments}
\begin{figure}[t]
    \centering
    \includegraphics[width=\linewidth]{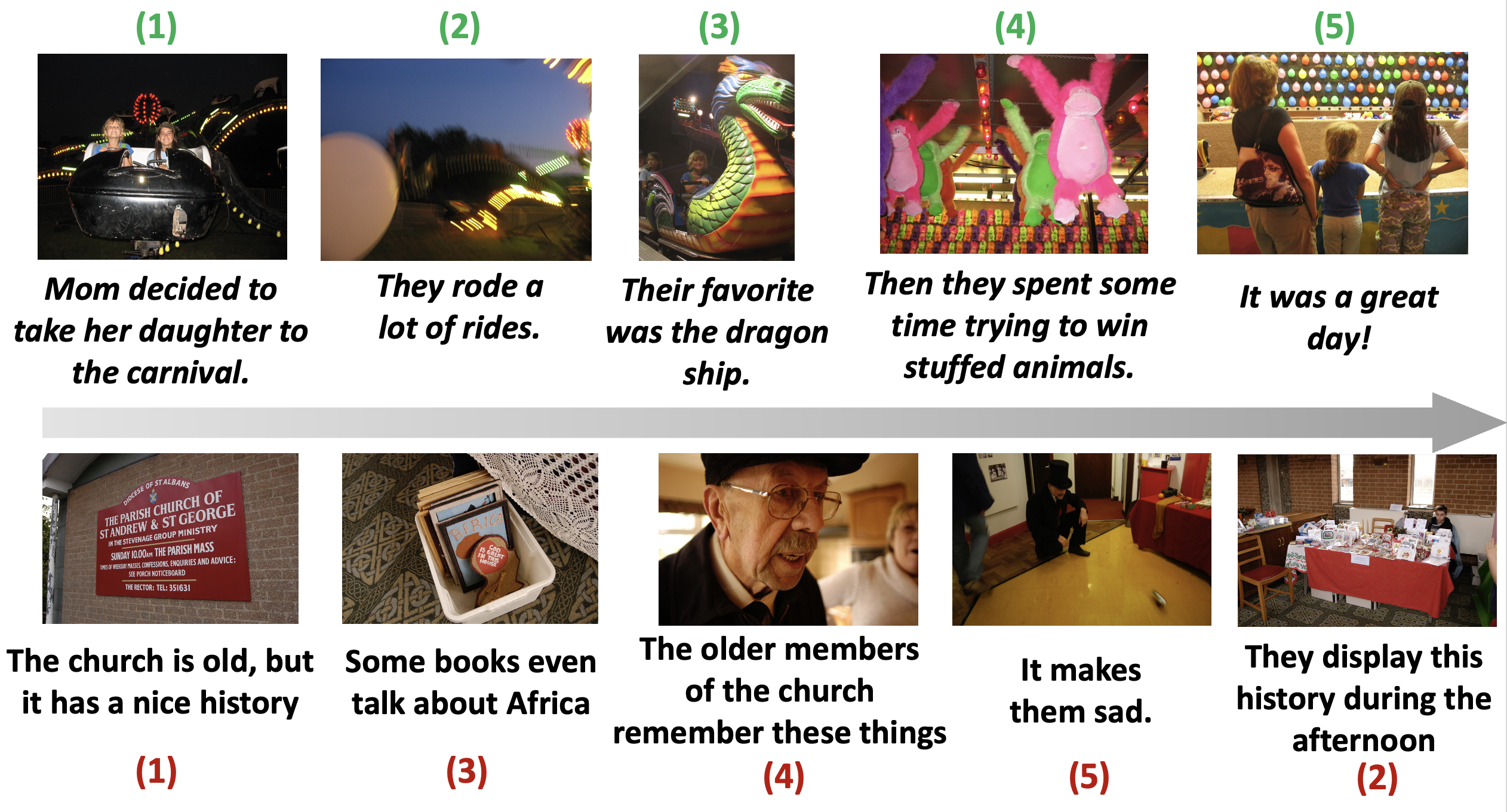}
    \caption{Successful (top) and failure (bottom) cases for ordering of a visual story from VIST~\cite{huang2016visual}. The image-sentence pairs are ordered from left to right as predicted by \mname. The numbers indicate the ground-truth positions of the image-caption in the story.}%
    \label{fig:vist}
    \vspace{-12pt}
\end{figure}

We finally evaluate \mname on a multi-modal task: visual storytelling,
which is harder than text reasoning, as it requires understanding and relating visual and text inputs. We follow the evaluation protocol in \cite{agrawal2016sort} using the visual storytelling dataset VIST \cite{huang2016visual}. The dataset contains stories of sentence-image pairs with each story describing everyday events, as shown in Fig.~\ref{fig:vist}.
\\
\noindent \textbf{Evaluation metrics.} We adopt three complementary metrics for the evaluation as in \cite{agrawal2016sort}:
\begin{itemize}[noitemsep,nolistsep,leftmargin=*]
\item \textit{Spearman's rank correlation (Sp.)} evaluates the monotonic relationship between predicted and ground-truth rankings. A higher score indicates better performance. 
\item \textit{Pairwise accuracy (Pairw.)} measures the percentage of pairs of elements with identical predicted and true ordering. A higher score indicates better performance.
\item \textit{Average Distance (Dist.)} measures the average displacement of all predicted elements to the ground truth order. A lower value indicates better performance.
\end{itemize}

\noindent \textbf{Implementation Details.} 
In this task, we are given five images with their related caption that have to be arranged correctly to form a visual story.
Similarly to Sentence Ordering, we divide a 1D space of range (-1,1) into five segments and assign to each image/text pair the center of its corresponding segment as the ground-truth position. The images/text at the beginning of the sequence have a lower position, while those at the end of the sequence have a larger position.
For this task, we extract the image feature using EfficientNet and extract a text summary for the phrase using BART as done for the Sentence ordering task. We then concatenate the output of these two models to obtain the features $\textbf{h}^i$. We train \mname with $T=100$ and sample it with inference ratio $r=10$.
\\
\noindent \textbf{Comparisons.} We compare \mname with \textit{Sort Story} as proposed by Agrawal \etal \cite{agrawal2016sort}. \textit{Sort Story} is a combination of Skip-Thought \cite{kiros2015skip} with pairwise ranking \cite{li2011short}, a custom CNN, and Neural Position Embedding module with LSTM \cite{graves2012long}. Agrawal \etal proposed two versions of their approach: an ensemble that adopts the Hungarian algorithm to find the permutation that maximizes the ensemble's voting and a standalone module. We also compare against a Transformer-based baseline with BART \cite{lewis2019bart} and EfficientNet \cite{tan2019efficientnet} for the textual and visual encoding, respectively.

Tab.~\ref{tab:results_vist} shows that \mname outperforms all the baselines in terms of the average distance (\textit{Dist.}). Our approach builds coherent stories with fewer wrong displacements compared to the ground truth. \mname also performs close to \textit{Sort Story}~\cite{agrawal2016sort} in terms of pairwise distance (\textit{pairw.}) and correlation (\textit{sp.}), which measure the global coherence, with only a slight difference. 
Note that \textit{Sort Story} adopts specific design choices including the pairwise distance between each visual-text pair, an ensemble of modules, and Hungarian matching algorithm, while our method is solely data-driven and relies on the reversing diffusion process. Fig.~\ref{fig:vist} shows some qualitative results. Our approach produces a coherent story even where the story's structure is loose (Fig.~\ref{fig:vist} top). These results confirm the advantage of our GNN formulation which enables each node to be coherent with others. In the failure case in Fig.~\ref{fig:vist} (bottom), \mname correctly predicts \textit{(1)} as first and \textit{(3)}-\textit{(4)}-\textit{(5)} in order, while misplaces \textit{(2)} as the last element of the story. %
As a plug-and-play method, \mname demonstrates competitive performance in building a coherent narration with multi-modal data.
\begin{table}[t]
\begin{center}
\resizebox{\linewidth}{!}{
\begin{tabular}{l|ccc}
\toprule
\textsc{Method} & \textbf{Sp.} $\uparrow$ & \textbf{Pairw.} $\uparrow$ & \textbf{Dist.} $\downarrow$ \\ 
\midrule
Sort Story$^\ddagger$ \cite{agrawal2016sort} & \textbf{0.67}& \textbf{0.79} & 0.72 \\
Skip-Thought + pairwise \cite{agrawal2016sort} & 0.56 & 0.74 & 0.89 \\
Transformer~\cite{vaswani2017attention} & 0.54 & 0.75 & \underline{0.56} \\
\hdashline
\textbf{\mname} & \underline{0.63} & \underline{0.77} & \textbf{0.51} \\
\bottomrule
\end{tabular}
}
\end{center}
\caption{Performance of different methods on the visual storytelling task. \textbf{Best} / \underline{Second best}. $\ddagger$ Ensemble of multiple methods.}
\label{tab:results_vist}
\end{table}

\section{Conclusion}
\label{sec:conclusion}

In this work, we proposed \mname, a graph-based DPM for positional reasoning on unordered sets. \mname represents the set as a fully connected graph where each element is a node of the graph. By using an Attention-based GNN, we update the node features to estimate the node position. The diffusion formulation allows us to learn the underlying patterns and iteratively refine the element positions.
\mname is generic and applicable to multiple tasks that require positional reasoning regardless of the data modality and positional dimension, as demonstrated in the experimental section. We experimented with three ordering tasks: puzzle solving, sentence ordering, and visual storytelling. \mname reaches SOTA on puzzle solving and comparable results with sentence ordering and visual storytelling when compared to methods that are specifically designed for each task.\\
\\
\textbf{Acknowledgments} We would like to thank Pietro Morerio, Davide Talon, and Theodore Tsesmelis for helping improving the manuscript and for the useful discussions.

{\small
\bibliographystyle{ieee_fullname}
\bibliography{egbib}

\begin{thebibliography}{10}\itemsep=-1pt

\bibitem{agrawal2016sort}
Harsh Agrawal, Arjun Chandrasekaran, Dhruv Batra, Devi Parikh, and Mohit
  Bansal.
\newblock Sort story: Sorting jumbled images and captions into stories.
\newblock {\em arXiv preprint arXiv:1606.07493}, 2016.

\bibitem{barzilay2008modeling}
Regina Barzilay and Mirella Lapata.
\newblock Modeling local coherence: An entity-based approach.
\newblock In {\em Proceedings of the 43rd Annual Meeting of the Association for
  Computational Linguistics ({ACL}{'}05)}, pages 141--148, Ann Arbor, Michigan,
  2005. Association for Computational Linguistics.

\bibitem{chen2016neural}
Xinchi Chen, Xipeng Qiu, and Xuanjing Huang.
\newblock Neural sentence ordering.
\newblock {\em arXiv preprint arXiv:1607.06952}, 2016.

\bibitem{choCVPR10probjigsaw}
T.~S. {Cho}, S. {Avidan}, and W.~T. {Freeman}.
\newblock A probabilistic image jigsaw puzzle solver.
\newblock In {\em IEEE Computer Society Conference on Computer Vision and
  Pattern Recognition}, pages 183--190, 2010.

\bibitem{chowdhury2021everything}
Somnath Basu~Roy Chowdhury, Faeze Brahman, and Snigdha Chaturvedi.
\newblock Is everything in order? a simple way to order sentences.
\newblock {\em arXiv preprint arXiv:2104.07064}, 2021.

\bibitem{cui2018deep}
Baiyun Cui, Yingming Li, Ming Chen, and Zhongfei Zhang.
\newblock Deep attentive sentence ordering network.
\newblock In {\em Proceedings of the 2018 Conference on Empirical Methods in
  Natural Language Processing}, pages 4340--4349, Brussels, Belgium, 2018.
  Association for Computational Linguistics.

\bibitem{cui2020bert}
Baiyun Cui, Yingming Li, and Zhongfei Zhang.
\newblock {BERT}-enhanced relational sentence ordering network.
\newblock In {\em Proceedings of the 2020 Conference on Empirical Methods in
  Natural Language Processing (EMNLP)}, pages 6310--6320, Online, 2020.
  Association for Computational Linguistics.

\bibitem{dhariwal2021diffusion}
Prafulla Dhariwal and Alexander Nichol.
\newblock Diffusion models beat gans on image synthesis.
\newblock {\em Advances in Neural Information Processing Systems}, 34, 2021.

\bibitem{elsner2011extending}
Micha Elsner and Eugene Charniak.
\newblock Extending the entity grid with entity-specific features.
\newblock In {\em Proceedings of the 49th Annual Meeting of the Association for
  Computational Linguistics: Human Language Technologies}, pages 125--129,
  2011.

\bibitem{gallagher2012jigsaw}
Andrew~C Gallagher.
\newblock Jigsaw puzzles with pieces of unknown orientation.
\newblock In {\em 2012 IEEE Conference on computer vision and pattern
  recognition}, pages 382--389. IEEE, 2012.

\bibitem{gong2016end}
Jingjing Gong, Xinchi Chen, Xipeng Qiu, and Xuanjing Huang.
\newblock End-to-end neural sentence ordering using pointer network.
\newblock {\em arXiv preprint arXiv:1611.04953}, 2016.

\bibitem{goodfellow2020generative}
Ian Goodfellow, Jean Pouget-Abadie, Mehdi Mirza, Bing Xu, David Warde-Farley,
  Sherjil Ozair, Aaron Courville, and Yoshua Bengio.
\newblock Generative adversarial networks.
\newblock {\em Communications of the ACM}, 63(11):139--144, 2020.

\bibitem{graves2012long}
Alex Graves and Alex Graves.
\newblock Long short-term memory.
\newblock {\em Supervised sequence labelling with recurrent neural networks},
  pages 37--45, 2012.

\bibitem{guinaudeau2013graph}
Camille Guinaudeau and Michael Strube.
\newblock Graph-based local coherence modeling.
\newblock In {\em Proceedings of the 51st Annual Meeting of the Association for
  Computational Linguistics (Volume 1: Long Papers)}, pages 93--103, 2013.

\bibitem{ho2020denoising}
Jonathan Ho, Ajay Jain, and Pieter Abbeel.
\newblock Denoising diffusion probabilistic models.
\newblock {\em Advances in Neural Information Processing Systems},
  33:6840--6851, 2020.

\bibitem{hoogeboom2022equivariant}
Emiel Hoogeboom, V{\i}ctor~Garcia Satorras, Cl{\'e}ment Vignac, and Max
  Welling.
\newblock Equivariant diffusion for molecule generation in 3d.
\newblock In {\em International Conference on Machine Learning}, pages
  8867--8887. PMLR, 2022.

\bibitem{huang2016visual}
Ting-Hao~Kenneth Huang, Francis Ferraro, Nasrin Mostafazadeh, Ishan Misra,
  Aishwarya Agrawal, Jacob Devlin, Ross Girshick, Xiaodong He, Pushmeet Kohli,
  Dhruv Batra, C.~Lawrence Zitnick, Devi Parikh, Lucy Vanderwende, Michel
  Galley, and Margaret Mitchell.
\newblock Visual storytelling.
\newblock In {\em Proceedings of the 2016 Conference of the North {A}merican
  Chapter of the Association for Computational Linguistics: Human Language
  Technologies}, pages 1233--1239, San Diego, California, 2016. Association for
  Computational Linguistics.

\bibitem{kiros2015skip}
Ryan Kiros, Yukun Zhu, Russ~R Salakhutdinov, Richard Zemel, Raquel Urtasun,
  Antonio Torralba, and Sanja Fidler.
\newblock Skip-thought vectors.
\newblock {\em Advances in neural information processing systems}, 28, 2015.

\bibitem{kumar2020deep}
Pawan Kumar, Dhanajit Brahma, Harish Karnick, and Piyush Rai.
\newblock Deep attentive ranking networks for learning to order sentences.
\newblock In {\em The Thirty-Fourth {AAAI} Conference on Artificial
  Intelligence, {AAAI} 2020, The Thirty-Second Innovative Applications of
  Artificial Intelligence Conference, {IAAI} 2020, The Tenth {AAAI} Symposium
  on Educational Advances in Artificial Intelligence, {EAAI} 2020, New York,
  NY, USA, February 7-12, 2020}, pages 8115--8122. {AAAI} Press, 2020.

\bibitem{LapataB05}
Mirella Lapata and Regina Barzilay.
\newblock Automatic evaluation of text coherence: Models and representations.
\newblock In Leslie~Pack Kaelbling and Alessandro Saffiotti, editors, {\em
  IJCAI-05, Proceedings of the Nineteenth International Joint Conference on
  Artificial Intelligence, Edinburgh, Scotland, UK, July 30 - August 5, 2005},
  pages 1085--1090. Professional Book Center, 2005.

\bibitem{CelebAMask-HQ}
Cheng-Han Lee, Ziwei Liu, Lingyun Wu, and Ping Luo.
\newblock Maskgan: Towards diverse and interactive facial image manipulation.
\newblock In {\em IEEE Conference on Computer Vision and Pattern Recognition
  (CVPR)}, 2020.

\bibitem{levine2012early}
Susan~C Levine, Kristin~R Ratliff, Janellen Huttenlocher, and Joanna Cannon.
\newblock Early puzzle play: a predictor of preschoolers' spatial
  transformation skill.
\newblock {\em Developmental psychology}, 48(2):530, 2012.

\bibitem{lewis2019bart}
Mike Lewis, Yinhan Liu, Naman Goyal, Marjan Ghazvininejad, Abdelrahman Mohamed,
  Omer Levy, Veselin Stoyanov, and Luke Zettlemoyer.
\newblock {BART}: Denoising sequence-to-sequence pre-training for natural
  language generation, translation, and comprehension.
\newblock In {\em Proceedings of the 58th Annual Meeting of the Association for
  Computational Linguistics}, pages 7871--7880, Online, 2020. Association for
  Computational Linguistics.

\bibitem{li2011short}
Hang Li.
\newblock A short introduction to learning to rank.
\newblock {\em IEICE TRANSACTIONS on Information and Systems},
  94(10):1854--1862, 2011.

\bibitem{logeswaran2018sentence}
Lajanugen Logeswaran, Honglak Lee, and Dragomir~R. Radev.
\newblock Sentence ordering and coherence modeling using recurrent neural
  networks.
\newblock In Sheila~A. McIlraith and Kilian~Q. Weinberger, editors, {\em
  Proceedings of the Thirty-Second {AAAI} Conference on Artificial
  Intelligence, (AAAI-18), the 30th innovative Applications of Artificial
  Intelligence (IAAI-18), and the 8th {AAAI} Symposium on Educational Advances
  in Artificial Intelligence (EAAI-18), New Orleans, Louisiana, USA, February
  2-7, 2018}, pages 5285--5292. {AAAI} Press, 2018.

\bibitem{luo2021diffusion}
Shitong Luo and Wei Hu.
\newblock Diffusion probabilistic models for 3d point cloud generation.
\newblock In {\em Proceedings of the IEEE/CVF Conference on Computer Vision and
  Pattern Recognition}, pages 2837--2845, 2021.

\bibitem{oh2019topic}
Byungkook Oh, Seungmin Seo, Cheolheon Shin, Eunju Jo, and Kyong-Ho Lee.
\newblock Topic-guided coherence modeling for sentence ordering by preserving
  global and local information.
\newblock In {\em Proceedings of the 2019 Conference on Empirical Methods in
  Natural Language Processing and the 9th International Joint Conference on
  Natural Language Processing (EMNLP-IJCNLP)}, pages 2273--2283, Hong Kong,
  China, 2019. Association for Computational Linguistics.

\bibitem{PaikinCVPR2015}
Genady Paikin and Ayellet Tal.
\newblock Solving multiple square jigsaw puzzles with missing pieces.
\newblock {\em IEEE CVPR}, pages 4832--4839, 2015.

\bibitem{pomeranzCVPR11greedy}
D. {Pomeranz}, M. {Shemesh}, and O. {Ben-Shahar}.
\newblock A fully automated greedy square jigsaw puzzle solver.
\newblock In {\em CVPR 2011}, pages 9--16, 2011.

\bibitem{prabhumoye2020topological}
Shrimai Prabhumoye, Ruslan Salakhutdinov, and Alan~W Black.
\newblock Topological sort for sentence ordering.
\newblock In {\em Proceedings of the 58th Annual Meeting of the Association for
  Computational Linguistics}, pages 2783--2792, Online, 2020. Association for
  Computational Linguistics.

\bibitem{shi-graphtransformer}
Yunsheng Shi, Zhengjie Huang, Shikun Feng, Hui Zhong, Wenjing Wang, and Yu Sun.
\newblock Masked label prediction: Unified message passing model for
  semi-supervised classification.
\newblock In {\em Proceedings of the International Joint Conference on
  Artificial Intelligence (IJCAI)}, 2021.

\bibitem{sohl2015deep}
Jascha Sohl-Dickstein, Eric Weiss, Niru Maheswaranathan, and Surya Ganguli.
\newblock Deep unsupervised learning using nonequilibrium thermodynamics.
\newblock In {\em International Conference on Machine Learning}, pages
  2256--2265. PMLR, 2015.

\bibitem{song2020denoising}
Jiaming Song, Chenlin Meng, and Stefano Ermon.
\newblock Denoising diffusion implicit models.
\newblock {\em arXiv preprint arXiv:2010.02502}, 2020.

\bibitem{song2020score}
Yang Song, Jascha Sohl-Dickstein, Diederik~P Kingma, Abhishek Kumar, Stefano
  Ermon, and Ben Poole.
\newblock Score-based generative modeling through stochastic differential
  equations.
\newblock {\em arXiv preprint arXiv:2011.13456}, 2020.

\bibitem{talon2022ganzzle}
Davide Talon, Alessio Del~Bue, and Stuart James.
\newblock Ganzzle: Reframing jigsaw puzzle solving as a retrieval task using a
  generative mental image.
\newblock In {\em 2022 IEEE International Conference on Image Processing
  (ICIP)}, pages 4083--4087. IEEE, 2022.

\bibitem{tan2019efficientnet}
Mingxing Tan and Quoc Le.
\newblock Efficientnet: Rethinking model scaling for convolutional neural
  networks.
\newblock In {\em International conference on machine learning}, pages
  6105--6114. PMLR, 2019.

\bibitem{artgan2018}
Wei~Ren Tan, Chee~Seng Chan, Hernan Aguirre, and Kiyoshi Tanaka.
\newblock Improved artgan for conditional synthesis of natural image and
  artwork.
\newblock {\em IEEE Transactions on Image Processing}, 28(1):394--409, 2019.

\bibitem{vaswani2017attention}
Ashish Vaswani, Noam Shazeer, Niki Parmar, Jakob Uszkoreit, Llion Jones,
  Aidan~N Gomez, {\L}ukasz Kaiser, and Illia Polosukhin.
\newblock Attention is all you need.
\newblock {\em Advances in neural information processing systems}, 30, 2017.

\bibitem{verdine2014deconstructing}
Brian~N Verdine, Roberta~M Golinkoff, Kathryn Hirsh-Pasek, Nora~S Newcombe,
  Andrew~T Filipowicz, and Alicia Chang.
\newblock Deconstructing building blocks: Preschoolers' spatial assembly
  performance relates to early mathematical skills.
\newblock {\em Child development}, 85(3):1062--1076, 2014.

\bibitem{vinyals2015pointer}
Oriol Vinyals, Meire Fortunato, and Navdeep Jaitly.
\newblock Pointer networks.
\newblock In Corinna Cortes, Neil~D. Lawrence, Daniel~D. Lee, Masashi Sugiyama,
  and Roman Garnett, editors, {\em Advances in Neural Information Processing
  Systems 28: Annual Conference on Neural Information Processing Systems 2015,
  December 7-12, 2015, Montreal, Quebec, Canada}, pages 2692--2700, 2015.

\bibitem{wang2019hierarchical}
Tianming Wang and Xiaojun Wan.
\newblock Hierarchical attention networks for sentence ordering.
\newblock In {\em Proceedings of the AAAI Conference on Artificial
  Intelligence}, number~01, pages 7184--7191. {AAAI} Press, 2019.

\bibitem{yin2020enhancing}
Yongjing Yin, Fandong Meng, Jinsong Su, Yubin Ge, Lingeng Song, Jie Zhou, and
  Jiebo Luo.
\newblock Enhancing pointer network for sentence ordering with pairwise
  ordering predictions.
\newblock In {\em Proceedings of the AAAI Conference on Artificial
  Intelligence}, volume~34, pages 9482--9489. {AAAI} Press, 2020.

\bibitem{yin2019graph}
Yongjing Yin, Linfeng Song, Jinsong Su, Jiali Zeng, Chulun Zhou, and Jiebo Luo.
\newblock Graph-based neural sentence ordering.
\newblock In Sarit Kraus, editor, {\em Proceedings of the Twenty-Eighth
  International Joint Conference on Artificial Intelligence, {IJCAI} 2019,
  Macao, China, August 10-16, 2019}, pages 5387--5393. ijcai.org, 2019.

\bibitem{zellers2021merlot}
Rowan Zellers, Ximing Lu, Jack Hessel, Youngjae Yu, Jae~Sung Park, Jize Cao,
  Ali Farhadi, and Yejin Choi.
\newblock Merlot: Multimodal neural script knowledge models.
\newblock {\em Advances in Neural Information Processing Systems},
  34:23634--23651, 2021.

\bibitem{zhanglearning}
Yan Zhang, Jonathon Hare, and Adam Pr{\"u}gel-Bennett.
\newblock Learning representations of sets through optimized permutations.
\newblock In {\em International Conference on Learning Representations}, 2019.

\bibitem{zhang2018ICLR}
Yan Zhang, Jonathon Hare, and Adam Prügel-Bennett.
\newblock Learning representations of sets through optimized permutations.
\newblock In {\em ICLR}, 2019.

\bibitem{zhu2021neural}
Yutao Zhu, Kun Zhou, Jian-Yun Nie, Shengchao Liu, and Zhicheng Dou.
\newblock Neural sentence ordering based on constraint graphs.
\newblock In {\em Proceedings of the AAAI Conference on Artificial
  Intelligence}, volume~35, pages 14656--14664, 2021.

\end{thebibliography}
}
\end{document}